\def\year{2022}\relax
\documentclass[letterpaper]{article} 
\usepackage{aaai22}  
\usepackage{times}  
\usepackage{helvet}  
\usepackage{courier}  
\usepackage[hyphens]{url}  
\usepackage{graphicx} 
\usepackage{natbib}  
\usepackage{caption} 
\DeclareCaptionStyle{ruled}{labelfont=normalfont,labelsep=colon,strut=off} 
\usepackage{algorithm}
\usepackage{algorithmic}

\usepackage{amsmath,amsfonts,bm}

\def\eqref#1{equation~\ref{#1}}

\def\1{\bm{1}}

\def\vb{{\bm{b}}}

\def\vw{{\bm{w}}}
\def\vx{{\bm{x}}}
\def\vy{{\bm{y}}}

\def\mB{{\bm{B}}}

\def\mW{{\bm{W}}}

\DeclareMathAlphabet{\mathsfit}{\encodingdefault}{\sfdefault}{m}{sl}
\SetMathAlphabet{\mathsfit}{bold}{\encodingdefault}{\sfdefault}{bx}{n}

\def\sO{{\mathbb{O}}}

\newcommand{\R}{\mathbb{R}}

\DeclareMathOperator*{\argmin}{arg\,min}

\usepackage{lipsum}
\usepackage{comment}
\usepackage{multirow}
\usepackage{wrapfig}
\usepackage{algorithm}
\usepackage{algorithmic}
\usepackage{subfigure}

\usepackage{graphicx}
\usepackage{color}

\usepackage[english]{babel}
\usepackage{amsthm}

\newcommand{\st}{{\mbox{s.t.}}}

\newcommand{\D}{\mathcal{D}}

\newcommand{\T}{\mathcal{T}}

\newcommand{\balpha}{{\bm{\alpha}}}
\newcommand{\bbeta}{{\bm{\beta}}}

\def\eg{\emph{e.g.}} 
\def\ie{\emph{i.e.}}

\def\wrt{w.r.t.}

\usepackage{comment}

\usepackage{newfloat}
\usepackage{listings}
\floatstyle{ruled}
\newfloat{listing}{tb}{lst}{}
\floatname{listing}{Listing}
\title{Stretchable Cells Help DARTS Search Better}

\author {
    \small
    Tao Huang\textsuperscript{\rm 1}\quad
    Shan You\textsuperscript{\rm 1,2}\thanks{\footnotesize{Correspondence to: Shan You $<$\texttt{youshan@sensetime.com}$>$.}}\quad
    Yibo Yang\textsuperscript{\rm 3}\quad
    Zhuozhuo Tu\textsuperscript{\rm 4}\quad
    Fei Wang\textsuperscript{\rm 1}\quad
    Chen Qian\textsuperscript{\rm 1}\quad
    Changshui Zhang\textsuperscript{\rm 2}
}
\affiliations {
    \small
    \textsuperscript{\rm 1}SenseTime Research\quad
    \textsuperscript{\rm 2}Department of Automation, Tsinghua University,\\ Institute for Artiﬁcial Intelligence, Tsinghua University (THUAI),\\ Beijing National Research Center for Information Science and Technology (BNRist)\\
    \textsuperscript{\rm 3}Key Laboratory of Machine Perception (MOE), School of EECS, Peking University\\
    \textsuperscript{\rm 4}UBTECH Sydney AI Centre, School of Computer Science, The University of Sydney, Australia
    
}

\usepackage{bibentry}

\begin{document}

\maketitle

\begin{abstract}
Differentiable neural architecture search (DARTS) has gained much success in discovering flexible and diverse cell types. To reduce the evaluation gap, the supernet is expected to have identical layes with the target network. 
However, even for this consistent search, the searched cells often suffer from poor performance, especially for the supernet with fewer layers, as current DARTS methods are prone to wide and shallow cells, and this topology collapse induces sub-optimal searched cells. In this paper, we alleviate this issue by endowing the cells with explicit stretchability, so the search can be directly implemented on our stretchable cells for both operation type and topology simultaneously.   
Concretely, we introduce a set of topological variables and a combinatorial probabilistic distribution to explicitly model the target topology. With more diverse and complex topologies, our method adapts well for various layer numbers.
Extensive experiments on CIFAR-10 and ImageNet show that our stretchable cells obtain better performance with fewer layers and parameters. For example, our method can improve DARTS by 0.28\% accuracy on CIFAR-10 dataset with 45\% parameters reduced or 2.9\% with similar FLOPs on ImageNet dataset.
\end{abstract}

\section{Introduction}

Targeting at slipping the leash of human empirical limitations and liberating the manual efforts in designing networks, 
neural architecture search (NAS) emerges as a burgeoning tool to automatically seek promising network architectures in 
a data-driven manner. 
To accomplish the architecture search, early approaches mainly adopt reinforcement learning (RL)~\cite{BakerGNR17,ZophL17} 
or evolutionary algorithms~\cite{real2019regularized}. 
Nevertheless, it often involves hundreds of GPUs for computation and takes a large volume of GPU hours to finish the searching. 

For the sake of searching efficiency, pioneer work NASNet~\cite{nasnet} proposed to search on a cell level, where the searched cells can be stacked to develop task-specific networks. ~\cite{pham2018efficient,bender2018understanding} leverage weight-sharing scheme and amortize the training cost for each candidate architecture. Recently, DARTS~\cite{darts} makes the most of both sides and proposes a differentiable NAS variant. In DARTS, the search space is regarded as a supernet (i.e., graph), and all candidate architectures are derived as its sub-graphs. Besides, a set of operation variables are introduced to indicate the importance of different operations, and the optimal architecture corresponds to that with the largest importance. 

\begin{figure}[t]
	\centering
	\includegraphics[width=0.35\textwidth]{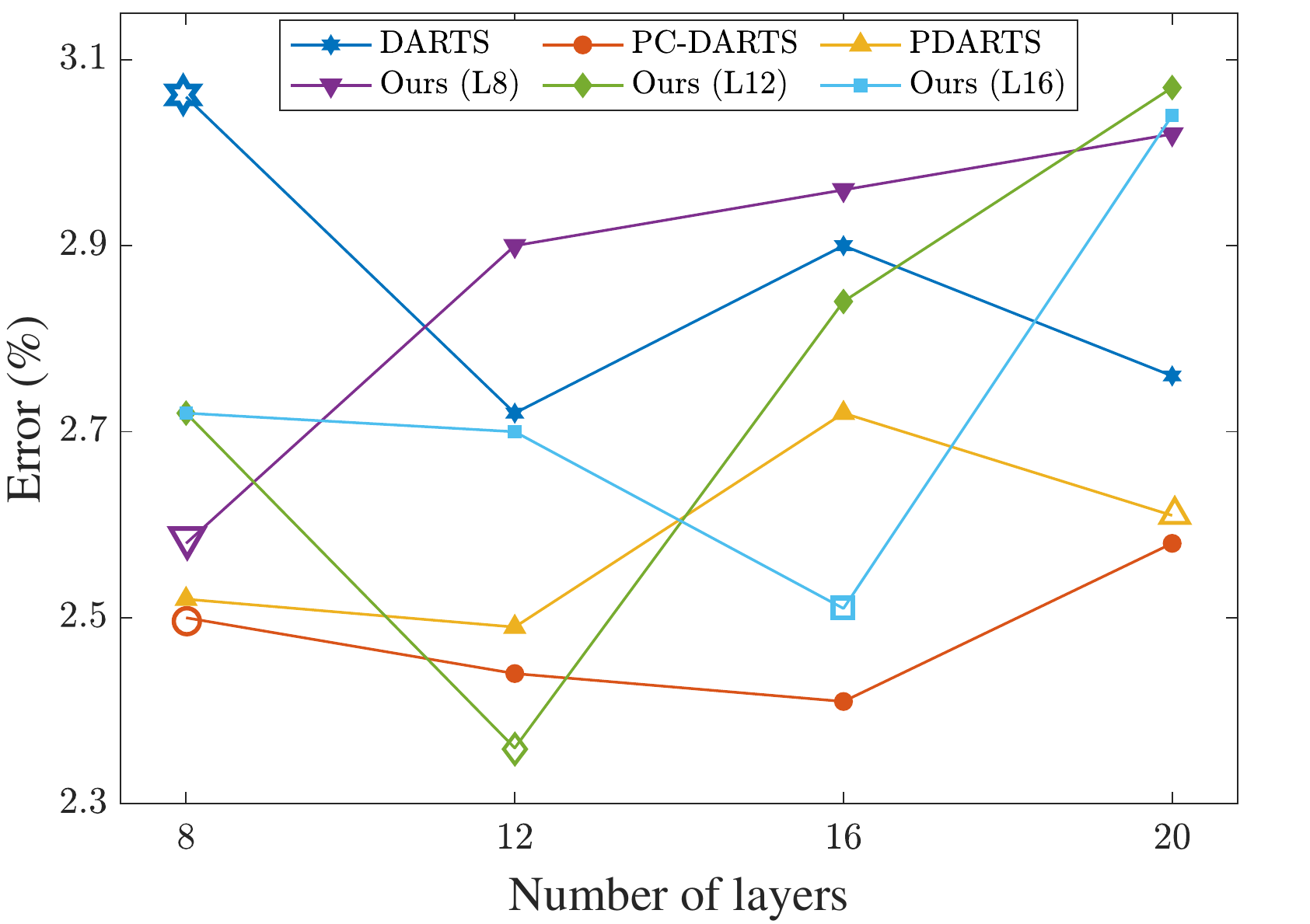}
	\caption{We train the searched cells of DARTS and its variants with different number of layers on CIFAR-10. Our stretchable cells obtain the lowest errors on the same numbers of layers in search, while other methods perform the best on different numbers. The hollow markers in figure indicate the same number of layers in search and evaluation.}
	\label{fig:layers}
	\vspace{-4mm}
\end{figure}

\begin{figure*}[t]
	\centering
	\includegraphics[width=0.9\textwidth,height=0.28\textwidth]{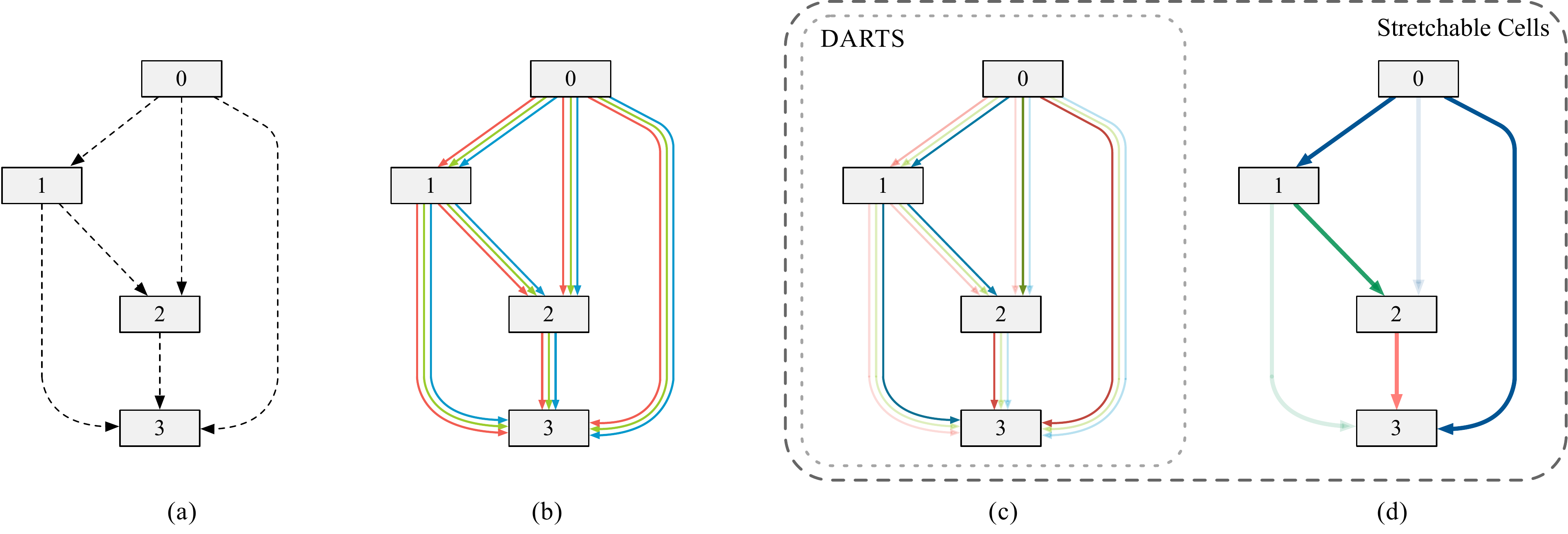}
	\vspace{-4mm}
	\caption{An overview of the proposed architecture modeling: (a) a cell represented by a directed acyclic graph. The edges between nodes denote the operations to be learned. (b) Following DARTS~\cite{darts}, the operation on each edge is replaced by a mixture of all candidate operations parameterized by operation variables $\balpha$. (c) DARTS selects operation with the largest $\balpha$ for each edge. (d) We introduce additional topological variables $\bbeta$ to explicitly learn topologies, which decouples operation selection (c) and topology learning (d).}
	\label{fig:overview}
	\vspace{-4mm}
\end{figure*}

Due to the limit of memory consumption, DARTS searches on an 8-layer supernet, and evaluates the searched cell by stacking it into a 20-layer target network. However, this induces an evaluation gap of different network depths; cells searched on an 8-layer supernet do not necessarily perform best on the target 20-layer network, and the evaluation of cells might be thus biased. Ideally, the search and the evaluation should be \textit{consistent}, \ie, the target network should be exactly the sub-network that is searched on the supernet. A common practice is to continue the training of the searched network in a one-stage manner \cite{chen2019progressive, bi2020gold} when the searching is finished. However, as shown in Figure~\ref{fig:layers}, we empirically observe that the derived cells of DARTS perform significantly poorly when also evaluated on an 8-layer target network, even for the cell in P-DARTS~\cite{chen2019progressive} searched with the increased layers, it still fails to obtain the best performance with the same number of layers ($20$), which is counterintuitive.

We argue this performance collapse results from the limited stretchability of cells; DARTS is prone to shallow and wide topologies~\cite{shu2019understanding}, such that the searched cells and the corresponding target network are way too shallow to have decent performance. In contrast, an ideal cell can adaptively stretch its topology during search according to the given supernet, and make its best to have appropriate topology and operations simultaneously. In this way, how to probe a good topology also matters for differentiable NAS. Nevertheless, DARTS and its variants~\cite{darts,pcdarts,milenas} currently couples the topology search into the operation search, and the topology is merely derived by a hand-crafted rule that keeps two edges for each node with the highest operation importance, which limits the learning of those architectures with truly-good topologies. 


In this paper, we propose a simple plug-and-play method to endow differentiable cells with explicit stretchability. Thus the NAS can be implemented directly on the stretchable cells for both optimal operation type and topology simultaneously. Concretely, as shown in Figure~\ref{fig:overview}, we decouple the modeling of operation and topology during search, and introduce a set of topological variables to indicate the topology learning. Instead of modeling each edge individually \cite{pcdarts}, we encourage each node to select edges freely with varying depths in the graph, and use the topological variables to model a combinatorial probabilistic distribution of all kinds of edge pairs; and the optimal topology corresponds to the edge pair with the largest topology score. 

The benefits of our method are in two folds. First, our stretchable cells can model sub-graphs with an arbitrary number of edges to search for the best depth, promoting the learned topology to be more diverse. Second, the proposed method can be applied to various differentiable NAS methods, and optimized jointly with operation variables and their weights as bi-level DARTS~\cite{darts} or mixed-level MiLeNAS~\cite{milenas}, which incurs almost no additional memory cost by merging the combinatorial probabilities as a factor. We also propose a \textit{passive-aggressive} regularization on the topological variables to eliminate invalid topology during the search. The empirical results on CIFAR-10 and ImageNet show that our method can help DARTS learn better cells with more diverse and complex topology corresponding to the number of layers. For example, within 1 GPU day of search cost, our stretchable cells can achieve 97.28\% accuracy on CIFAR-10 dataset but has only 1.8M parameters compared to 97.00\% accuracy with 3.3M parameters of DARTS. Meanwhile, with a similar amount of parameters, our method obtains 97.64\% accuracy with 3.5M parameters while the MiLeNAS merely has 97.49\% accuracy but with 3.9M parameters.

\section{Revisiting Differentiable NAS}
We first review the vanilla differentiable NAS method DARTS~\cite{darts}, which searchs for a computation cell as the building block of the final architecture. Mathematically, a cell can be considered as a directed acyclic graph (DAG) consisting of an ordered sequence of $N$ nodes. Each node $\vx_i$ is represented as a feature map, and each directed edge $(i,j)$ between nodes indicates the candidate operations $o\in\sO$, such as \texttt{max pooling}, \texttt{convolution} and \texttt{identity mapping}. Then the goal is to determine one operation $o$ from $\sO$ to connect each pair of nodes. DARTS relaxes this categorical operation selection into a soft and continuous selection using softmax probabilities with a set of variables $\balpha_{i,j}\in\R^{|\sO|}$ to indicate the operation importance, 
\begin{equation}
\label{eq:darts_ori}
o^{(i, j)}(\vx_i) = \sum_{o\in\sO} \frac{\exp(\alpha_{i,j}^o)}{\sum_{o'\in \sO} \exp(\alpha_{i,j}^{o'})} o(\vx_i),
\end{equation}
where $|\sO|$ is the number of all candidate operations, $o(\vx_i)$ is the result of applying operation $o$ on $\vx_i$, and $o^{(i, j)}(\vx_i)$ means the summed feature maps from $\vx_i$ to $\vx_j$. Then the output of a node $\vx_j$ is the sum of all feature maps from all its precedent nodes, with associated edges $\{(1,j), ..., (j-1,j)\}$, \ie, 
\begin{equation}
\label{eq:darts_node_ori}
\vx_j = \sum_{i<j}o^{i,j}(\vx_i) . 
\end{equation}
Moreover, each cell has two input nodes $\vx_1$ and $\vx_2$, which are the outputs of previous two cells, and the final output of an entire cell is formed by concatenating all intermediate nodes inside, \ie,  $\{\vx_3, ..., \vx_N\}$. The operation variables $\balpha$ can be jointly trained with supernet weights by gradient-based optimizers~\cite{darts,milenas}. After training, the optimal operation corresponds to the one with the maximum operation importance.

As for the derivation of topology, a hand-crafted rule is usually followed, which manually specify that only two input edges are active for each node $\vx_j$ by selecting the edges with the top-2 largest operation importance in $\{o^{(1, j)}, ..., o^{(j-1, j)}\}$. However, this heuristic rule fails to obtain more diverse topology, and thus does not find the optimal one. We will elaborate our proposed method in the sequel, which aims to explicitly learn stretchable cells.

\noindent\textbf{Remark.} The original DARTS introduces a \texttt{zero} operation for selecting edges; however, comparing different edges is not fair because they are calculated among operations on each edge independently, and it still needs to manually choose the two most important ones.
Meanwhile, this additional operation is believed to cause an optimization gap since the weight on \texttt{zero} operation is often very large (\eg, the weights of \texttt{zero} operations are usually larger than $0.7$ at the end of training in our experiments) compared to the small weights in other operations resulting in inaccurate comparison and a collapse problem.

\section{Proposed Approach: Stretchable Cells}
In this section, we formally elaborate our depth-aware topology learning method, which enables us to search stretchable cells and applies to various DARTS variants. 


\subsection{Decoupling topology and operations}
As previously illustrated, DARTS and its variants couple the operation selection and topology in their architecture modeling, and resort to a hand-crafted rule for deriving the topology. This practice usually induces sub-optimal results as the derived cells may not best suit the number of layers in search. Besides, the striking gap between search and evaluation also increases the uncertainty of performance. Both sides motivate us to highlight the learning of topology in differentiable NAS.

First recall that DARTS stipulates that only two input edges are allocated to each node. Now we investigate this \textit{fixed-topology} case. Instead of using the hand-crafted rule, we propose to learn automatically which two edges should be connected to each node. Based on the definition of DAG, the topology space $\T$ can be decomposed by each node and represented by their input edges, \ie, $\T = \bigotimes_{j=1}^N \tau_j$, where $\bigotimes$ is the Cartesian product of all $\tau_j$'s and $\tau_j$ is the set of all input edges pairs for node $\vx_j$, \ie, 
\begin{equation}\label{topo}
\tau_j = \{(1, 2), ..., (1, j-1), ..., (2, 3), ..., (j-2, j-1)\},
\end{equation}
and $|\tau_j| = \mathbb{C}_{j-1}^2$ since only two input edges are specified for each node. For the fixed input from the previous two cells, we denote them as $\vx_1$ and $\vx_2$, respectively, and we have $\tau_1=\tau_2 = \emptyset$.  

Since we have a clear and complete modeling over all possible topology $\T$ as Eq.(\ref{topo}), to determine the optimal topology, it is natural to introduce another set of topological variables $\bbeta=\{\bbeta_j\}_{j=1}^N$ with $|\bbeta_j| = |\tau_j|$, where $\bbeta_j$ explicitly represents the soft importance of each topology (\ie, input edge pairs). Then each node is rewritten as 
\begin{equation}\label{eq:topo_general}
\vx_j = \sum_{(m,n)\in\tau_j}  p_{j}^{(m, n)}  (o^{m,j}(\vx_m)+o^{n,j}(\vx_n)) ,
\end{equation}
where $p_{j}^{(m, n)}$ is the combinatorial probability for selecting edges $(m, j)$ and $(n, j)$ as $\vx_j$'s inputs, defined as
\begin{equation}
p_{j}^{(m, n)} = \frac{\exp(\beta_{j}^{(m, n)})}{\sum_{(m',n')\in\tau_j} \exp(\beta_{j}^{(m', n')})} ,
\end{equation}
and $\beta_{j}^{(m, n)}$ is the corresponding variable for $p_{j}^{(m, n)}$. Different from Eq.(\ref{eq:darts_node_ori}), Eq.(\ref{eq:topo_general}) considers a combinatorial probabilistic distribution over all possible valid topologies, and the optimal topology simply refers to the one with the largest topological importance. Note that since the topology and operations are already decoupled, there is no need to involve a \texttt{zero} operation. 

Directly computing Eq.(\ref{eq:topo_general}) will increase the memory consumption of feature maps since it needs to compute the summation of two feature maps $o^{m,j}(\vx_m)+o^{n,j}(\vx_n)$. Fortunately, by merging the combinatorial probabilities associated with the same edges, this issue can be well addressed and Eq.(\ref{eq:topo_general}) can be simplified as
\begin{align}\label{merge}
\begin{split}
&\vx_j = \sum_{i<j} s(i, j) \cdot o^{i,j}(\vx_i),\quad \\
\mbox{with}\quad &s(i,j) = \sum_{k<i} p_{j}^{(k,i)} + \sum_{i<k<j}  p_{j}^{(i, k)} ,
\end{split}
\end{align}
where $s(i,j)$ is the merged combinatorial probability \wrt~node $\vx_i$. In this way, the memory cost of Eq.(\ref{merge}) is almost the same as DARTS,  with only a learnable $s(i,j)$ added before edge accumulation.

\bigbreak
\noindent\textbf{Switching topology space with output edges\ }\\
According to the ablation study in our experiments, solving Eq.(\ref{merge}) leads to poor performance. This might because we examine the topology space from the perspective of input edges. In other words, we investigate which input edge pair should be selected to connect each node. However, as the \textit{skip-dominant issue}~\cite{pcdarts, liang2019darts, bi2019stabilizing} in operation selection, precedent features of a node will dominate its current features during the training of supernet. 
Similarly, edges from the more precedent nodes will also tend to be dominant since they can be regarded as edge-level skip layers and benefit more from optimization.
Since a DAG can be both described by the input edges and output edges, we propose to switch the topology space to the perspective of output edges. Selecting the output edges will alleviate the dominant issue on topology since features from the same node are compared.

Concretely, we examine the topology by selecting which edge pair should be the output for each node. By doing so, all output edges use the same nodes as input and can be compared more fairly. In this way, the output nodes of each node $\vx_i$ are from its posterior nodes $\vx_{i+1},...,\vx_{N}$, and we switch the topology space Eq.(\ref{topo}) to 
\begin{align}
\begin{split}
\tilde{\tau}_i = \{&(i+1,i+2), ..., (i+1, N), ..., \\
&(i+2, i+3), ..., (N-1, N)\} ,
\end{split}
\end{align}
which is the set of output edge pairs of node $\vx_i$ from all $N-i$ edges. Similar to the previous discussions, we also introduce a set of topological variables $\tilde{\bbeta}=\{\tilde{\bbeta}_i\}_{i=1}^N$ to model the soft importance for each edge pair, and $\tilde{p}_{i}^{(m, n)}$ denotes the combinatorial probability of choosing $(i, m)$ and $(i, n)$ as output edges of node $\vx_i$ with the corresponding variable $\tilde{\beta}_i^{(m,n)}$ and $|\tilde{\bbeta}_i| = |\tilde{\tau}_i| = \mathbb{C}_{N-i}^2$, 
\begin{equation}
\tilde{p}_{i}^{(m, n)} = \frac{\exp(\tilde{\beta}_i^{(m,n)})}{\sum_{(m',n')\in\tilde{\tau}_i} \exp(\tilde{\beta}_i^{(m',n')})}.
\end{equation}
Then a node is also represented as Eq.(\ref{merge}), but with different merged probabilities, \ie,  
\begin{align}\label{twoedge}
\begin{split}
&\vx_j = \sum_{i<j} \tilde{s}(i, j) \cdot o^{i,j}(\vx_i), \\
\mbox{with}\quad &\tilde{s}(i, j)= \sum_{i<k<j} \tilde{p}_{i}^{(k, j)}    +  \sum_{k>j} \tilde{p}_{i}^{(j,k)} . 
\end{split}
\end{align}
Since the number of posterior nodes for each node is not equal (\eg, node $\vx_2$ has 4 posterior nodes with $N=6$, but node $\vx_3$ only has 3) , the merged probabilities $\tilde{s}(i, j)$ associated with each edge $(i, j)$ have different magnitude of value. We scale it for more stable optimization, \ie, 
\begin{equation}
\tilde{s}(i, j) =  \frac{\mathbb{C}_{N-i}^2}{\mathbb{C}_{N-i-1}^1} (\sum_{i<k<j} \tilde{p}_{i}^{(k, j)}    +  \sum_{k>j} \tilde{p}_{i}^{(j,k)}) .
\end{equation}

\subsection{Generalizing to stretchable cells}
The previous section considers a fixed topology space, since we only allocate two output edges for each node and model a combinatorial probability distribution over all possible output edge pairs. This manually designed rule of edge selection can only learn a cell with a fixed number of operations, and the diversity of cell depth is limited. In this section, we show that our method can be naturally generalized to an arbitrary topology space to learn stretchable cells, if we do not restrict the number of output edges for each node and allow each node to connect its posterior nodes freely. 

For arbitrary topology modeling, each node $\vx_i$ can be connected with any number (including none for removing the node) of its posterior nodes $\vx_{i+1},...,\vx_{N}$, and thus the amount of all possible combinatorial pairs $\hat{\tau}_i$ becomes 
\begin{equation}
|\hat{\tau}_i| = \sum_{n=1}^{N-i} \mathbb{C}_{N-i}^n = 2^{N-i},
\end{equation}
and each combinatorial pair can be uniquely defined by a binary code vector, \ie, $\vb_i = (b^{i+1}_i, b^{i+2}_i, ..., b^{N}_i)$ with $b^k_i \in \{0, 1\}$, where $b^k_i=1$ if edge $(i, k)$ exists and $b^k_i=0$ otherwise. Let $\mB_i = \{\vb_i^{(1)}, ..., \vb_i^{(|\hat{\tau}_i|)}\}$ denotes the set of all valid binary code vectors for $\vx_i$, we impose a combinatorial probability distribution to indicate the importance for each topology (\ie, binary node vector), \ie, 
\begin{equation}
\hat{p}(\vb_i) = \frac{\exp(\beta_i^{\vb_i})}{\sum_{\vb'_{i} \in \mB_i}\exp(\beta_i^{\vb'_i})},
\end{equation}
where we denote $\beta_i^{\vb_i}\in\R$ as the introduced topology variable corresponding to the binary code $\vb_i$. Then Eq.(\ref{twoedge}) can be rewritten similarly, but also with a different merged probability for each node, \ie, 
\begin{equation}
\hat{s}(i,j) = \frac{2^{N-i} -1}{2^{N-i-1}} \sum_{\vb_i\in \mB_i, b_i^{j}=1} \hat{p}(\vb_i). 
\end{equation}

\subsection{Regularizing invalid topologies during search}
Though the topology can be explicitly modeled in our method, there might be invalid topologies during search if a node is not connected with any of its precedent nodes. To suppress this trivial case, we introduce a \textit{passive-aggressive} regularization during search, \ie, 
\begin{equation}
\label{eq:reg}
r(\bbeta) = \sum_{j<N} \prod_{i<j}(1 - \max_{\vb_i \in \mB_i, \vb_i^{j}=1}\hat{p}(\vb_i) / \max_{\vb_i \in \mB_i}\hat{p}(\vb_i)) ,
\end{equation}
where $\max_{\vb_i \in \mB_i}\hat{p}(\vb_i)$ denotes the max probability among all the output edges of node $\vx_i$, and $\max_{\vb_i \in \mB_i, \vb_i^{j}=1}\hat{p}(\vb_i)$ denotes the max probability associated to edge $(i,j)$ . 

Note that $r(\bbeta)$ only \textit{aggressively} punishes the topology variables which predict invalid topology, while it \textit{passively} does no harm to the optimization when the topology is valid. If edge $(i,j)$ is chosen to be kept in the final architecture, it will hold that $\max_{\vb_i \in \mB_i}\hat{p}(\vb_i) - \max_{\vb_i \in \mB_i, \vb_i^{j}=1}\hat{p}(\vb_i) = 0$, and $r(\bbeta)=0$ if the architecture is valid. This regularization can be integrated into the loss \wrt~$\bbeta$, \ie,
\begin{equation}
\mathcal{L}_{val_\bbeta}(\mW,\balpha,\bbeta) = \mathcal{L}_{task}(\hat{\vy}, \vy) + \lambda \cdot r(\bbeta),
\end{equation}
where $\mathcal{L}_{task}$ is task-specific loss, and we set $\lambda$ as $10$ in our experiments.

\begin{table*}
	\renewcommand\arraystretch{1.17}
	\setlength\tabcolsep{3mm}
	\centering
	\caption{Search results on CIFAR-10 and comparison with state-of-the-art methods. $^\dag$: reported by DARTS~\cite{darts}, $^\ddag$: we fix the topology as \textit{MiLeNAS-SC-L8} and search operations randomly using the same strategy as DARTS. Search cost is tested on single NVIDIA GTX 1080 Ti GPU.}
	\vspace{-2mm}
	\label{tab:cifar10}   
	\footnotesize
	\begin{tabular}{lccccc}
		\hline
		\multirow{2}*{Methods} & Test Error  & Params & FLOPs & Search Cost  & \multirow{2}*{Search Method}\\
		~  & (\%) &  (M) & (M) & (GPU days) & ~\\
		\hline
		DenseNet-BC~\cite{huang2017densely} & 3.46 & 25.6 & - & - & manual\\
		\hline
		NASNet-A + cutout~\cite{nasnet} & 2.65 & 3.3 & - & 1800 & RL\\
		ENAS + cutout~\cite{pham2018efficient} & 2.89 & 4.6 & - & 0.5  & RL\\
		AmoebaNet-B +cutout~\cite{real2019regularized} & 2.55$\pm$0.05 & 2.8 & - & 3150 & evolution \\
		\hline
		PC-DARTS + cutout~\cite{pcdarts} & 2.57$\pm$0.07 & 3.6 & 576 & 0.1 & gradient-based\\  
		P-DARTS + cutout~\cite{chen2019progressive} & 2.50 & 3.4 & 551 & 0.3 & gradient-based \\
		GOLD-NAS-L + cutout~\cite{bi2020gold} & 2.53$\pm$0.08 & 3.67 & 546 & 1.1 & gradient-based \\
		SI-VDNAS (base) + cutout~\cite{wang2020si} & 2.50$\pm$0.06 & 3.6 & - & 0.3 & gradient-based\\
		DARTS (1st order) + cutout~\cite{darts} & 3.00$\pm$0.14 & 3.3 & 519 & 0.4 & gradient-based \\
		DARTS (2nd order) + cutout~\cite{darts} & 2.76$\pm$0.09 & 3.3 & 547 & 4.0 & gradient-based \\
		MiLeNAS + cutout~\cite{milenas} & 2.51$\pm$0.11 & 3.87 & 629 & 0.3 & gradient-based\\
		\hline
		Random search baseline$^\dag$ + cutout & 3.29$\pm$0.15 & 3.2 & - & 4 & random \\ 
		MiLeNAS-SC-L8 (random operation)$^\ddag$ + cutout & 2.77$\pm$0.08 & 3.3 & 576 & 4 & random \\
		\hline
		DARTS w/o SC + cutout & 2.72$\pm$0.12 & \textbf{1.8} & 337 & 0.6 & gradient-based \\ 
		DARTS w/o SC + cutout (C=48) & 2.48$\pm$0.09& 3.2 & 578 & 0.6 & gradient-based \\ 
		\hline
        MiLeNAS-SC-L8 + cutout & 2.58$\pm$0.11 & 3.5 & 599 & 1.1 & gradient-based \\
        MiLeNAS-SC-L12 + cutout & 2.36$\pm$0.07 & 3.5 & 622 & 1.5 & gradient-based \\
        MiLeNAS-SC-L16 + cutout & 2.51$\pm$0.09 & 3.57 & 577 & 2.2 & gradient-based \\
		\hline
	\end{tabular}
	\vspace{-4mm}
\end{table*}
\subsection{Integrating stretchable cells into DARTS variants}
Now we illustrate how our proposed stretchable cells can be applied to DARTS variants to boost their performance. Details can be found in supplementary materials. 

\textbf{DARTS.\ } The original DARTS~\cite{darts} formulates the NAS into a bi-level optimization problem~\cite{anandalingam1992hierarchical, colson2007overview}:
\begin{align}\label{eq:darts}
\begin{split}
&\min_\balpha\ \ \mathcal{L}_{val}(\vw^*(\balpha), \balpha), \\
&\st \vw^*(\balpha) = \argmin_\vw \mathcal{L}_{train}(\vw, \balpha),
\end{split}
\end{align}
where the operation variables $\balpha$ and supernet weights $\vw$ can be jointly optimized. We introduce additional topology variables $\bbeta$ to DARTS and Eq.(\ref{eq:darts}) becomes
\begin{align}\label{eq:topo}
\begin{split}
&\min_{\balpha,\bbeta}\ \ \mathcal{L}_{val}(\vw^*(\balpha, \bbeta), \balpha, \bbeta), \\
&\st~ \vw^*(\balpha, \bbeta) = \argmin_\vw \mathcal{L}_{train}(\vw, \balpha, \bbeta).
\end{split}
\end{align}
Since solving Eq.(\ref{eq:topo}) involves additional topology optimization, for efficiency consideration, it suffices to adopt the first-order approximation in our experiments. 


\textbf{MiLeNAS.\ } MiLeNAS~\cite{milenas} proposes a mixed-level reformulation of DARTS, which aims to optimize NAS more efficiently and reliably by mixing the training and validation loss for architecture optimization. Analogously, the introduced topology variables $\bbeta$ can be applied to MiLeNAS, and the objective becomes,
\begin{align}\label{eq:mile}
\begin{split}
\min_{\balpha,\bbeta}\quad &\mathcal{L}_{tr}(\vw^*(\balpha, \bbeta),\balpha,\bbeta)\\ 
&+ \lambda' \cdot \mathcal{L}_{val}(\vw^*(\balpha, \bbeta), \balpha, \bbeta),\quad 
\end{split}
\end{align}
which can also be solved efficiently using the first-order approximation as MiLeNAS. 

\section{Experiments} \label{exp}

\subsection{Datasets and implementation details}
We perform experiments on two benchmark datasets CIFAR-10~\cite{cifar10} and ImageNet~\cite{Imagenet}. 
For fair comparison, the operation space $\sO$ in our method is similar to DARTS, 
the only difference is that we remove \texttt{zero} operation as our method can explicitly learn the topology. 

\textbf{Search without stretchable cells}\ \ 
We first compare our method with DARTS using a fixed number of edges (without stretchable cells), which keeps the same edge number $8$ as DARTS. 

\textbf{Search with stretchable cells}\ \ We further perform architecture search with stretchable cells. For a diverse cell depth, we increase the number of nodes $N$ to $9$. Every node in the cell can choose to connect to any (including none for removing the node) of its posterior nodes. Besides, we conduct search with multiple numbers of layers, including $8$, $12$, and $16$, and the layers in evaluation are the same as search.

Detailed experimental setup can be found in supplementary materials.

\begin{table*}
	\renewcommand\arraystretch{1.17}
	\setlength\tabcolsep{3mm}
	\centering
	\footnotesize
	\caption{Search results on ImageNet and comparison with state-of-the-art methods.}
	\label{tab:imagenet}
	\vspace{-2mm}
	\begin{tabular}{lcccccc}
		\hline
		\multirow{2}*{Methods} & \multicolumn{2}{c}{Test Err. (\%)} & Params & Flops & Search Cost & \multirow{2}*{Search Method}\\
		\cline{2-3}
		~ & top-1 & top-5 &  (M) & (M) & (GPU days) & ~\\
		\hline
		MobileNet~\cite{howard2017mobilenets} & 29.4 & 10.5 & 4.2 & 569 & - & manual\\
		ShuffleNetV2 2$\times$~\cite{ma2018shufflenet} & 25.1 & - & $\sim$5 & 591 & - & manual\\
		\hline
		MnasNet-92~\cite{tan2019mnasnet} & 25.2 & 8.0 & 4.4 & 388 & - & RL\\
		AmoebaNet-C~\cite{real2019regularized} & 24.3 & 7.6 & 6.4 & 570 & 3150 & evolution\\
		\hline
		DARTS (2nd order)~\cite{darts} & 26.7 & 8.7 & 4.7 & 574 & 4.0 & gradient-based\\
		SNAS~\cite{snas} & 27.3 & 9.2 & 4.3 & 522 & 1.5 & gradient-based\\
		ProxylessNAS (ImageNet)~\cite{proxylessnas} & 24.9 & 7.5 & 7.1 & 465 & 8.3 & gradient-based\\
		P-DARTS~\cite{chen2019progressive} & 24.4 & 7.4 & 4.9 & 557 & 0.3 & gradient-based\\
		HM-NAS~\cite{yan2019hm} & 26.6 & - & 3.6 & 482 & $\sim$2 & gradient-based \\
		PC-DARTS (CIFAR-10)~\cite{pcdarts} & 25.1 & 7.8 & 5.3 & 586 & 0.1 & gradient-based\\
		PC-DARTS (ImageNet)~\cite{pcdarts} & 24.2 & 7.3 & 5.3 & 597 & 3.8 & gradient-based\\
		GOLD-NAS-Z-tr~\cite{bi2020gold} & 23.9 & 7.3 & 6.4 & 590 & 1.7 & gradient-based\\
		SI-VDNAS (base)~\cite{wang2020si} & 25.3 & 8.0 & 5.0 & 577 & 0.3 & gradient-based\\
		MiLeNAS~\cite{milenas} & 24.7 & 7.6 & 5.3 & 584 & 0.3 & gradient-based \\
		\hline
		MiLeNAS-SC-L12 (CIFAR-10)& 24.2 & 7.5 & 5.3 & 582 & 1.5 & gradient-based \\
		MiLeNAS-SC-L8 (ImageNet)& \textbf{23.8} & 7.2 & 5.6 & 612 & 6.2 & gradient-based \\
		\hline
	\end{tabular}
	\vspace{-4mm}
\end{table*}

\subsection{Results on CIFAR-10 dataset}

As previously discussed, for fair comparisons, we first adopt fixed-edge search without stretchable cells using the first-order approximation in DARTS, and the obtained model is named \textit{DARTS w/o SC}. Then we conduct search with stretchable cells on multiple numbers of layers using the approximation in MiLeNAS. The obtained models with $l$ layers are named \textit{MiLeNAS-SC-L}$l$, where $l\in\{8, 12, 16\}$.

In the search stage, the supernet is built by stacking $8$ cells in \textit{DARTS w/o SC} or $l$ cells in \textit{MiLeNAS-SC-L}$l$. We directly choose the latest optimized networks after training $50$ epochs with batch size $64$ for deriving architectures.

Unlike DARTS using $20$ stacked cells to build evaluation networks, all our obtained networks use the same numbers of layers as in their search. Our evaluation results on CIFAR-10 dataset compared with recent approaches are summarized in Table \ref{tab:cifar10}. Our method can obtain competitive results but with much fewer parameters, for example, \textit{DARTS w/o SC} can achieve 2.72\% test error with only 1.8M parameters, which reduces $\sim45\%$ parameters with higher accuracy compared to the original DARTS. That might be because our stretchable cells adapt well on the layer number, and the cells with small layer numbers are more deeper and thinner compared to DARTS. Thus in evaluation, we can stack fewer cells to reduce parameters. Besides, for a fair comparison with other methods, we enlarge the initial channel number $C$ of our networks to increase their parameters and FLOPs. The results show that, with a similar amount of parameters, our stretchable cells can achieve lower test error, significantly outperforming our baseline methods DARTS and MiLeNAS. Meanwhile, with stretchable cell search, our method can further obtain higher performance with varying layer numbers.

Moreover, to show the importance of stretchable cells, we adopt the same random search strategy as DARTS by keeping the cell topology fixed as \textit{MiLeNAS-SC-L8} and search operations randomly, marked as \textit{MiLeNAS-SC-L8 (random operation)}. From the results we can infer that, topologies contribute a lot to the performance, and our method learns a good topology for the network. Note that our random baseline significantly outperforms \textit{DARTS (1st order)}, demonstrating that we should highlight the topology learning.

\begin{figure}[t]
	\centering
   \subfigure[MiLeNAS-SC-L8] {\includegraphics[width=0.8\linewidth]{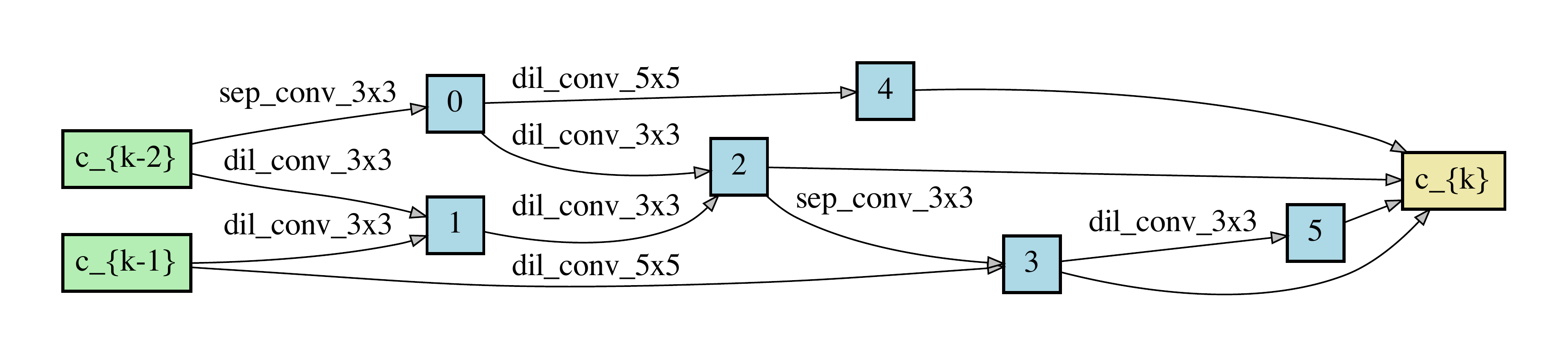}}
   \subfigure[MiLeNAS-SC-L16] {\includegraphics[width=0.65\linewidth]{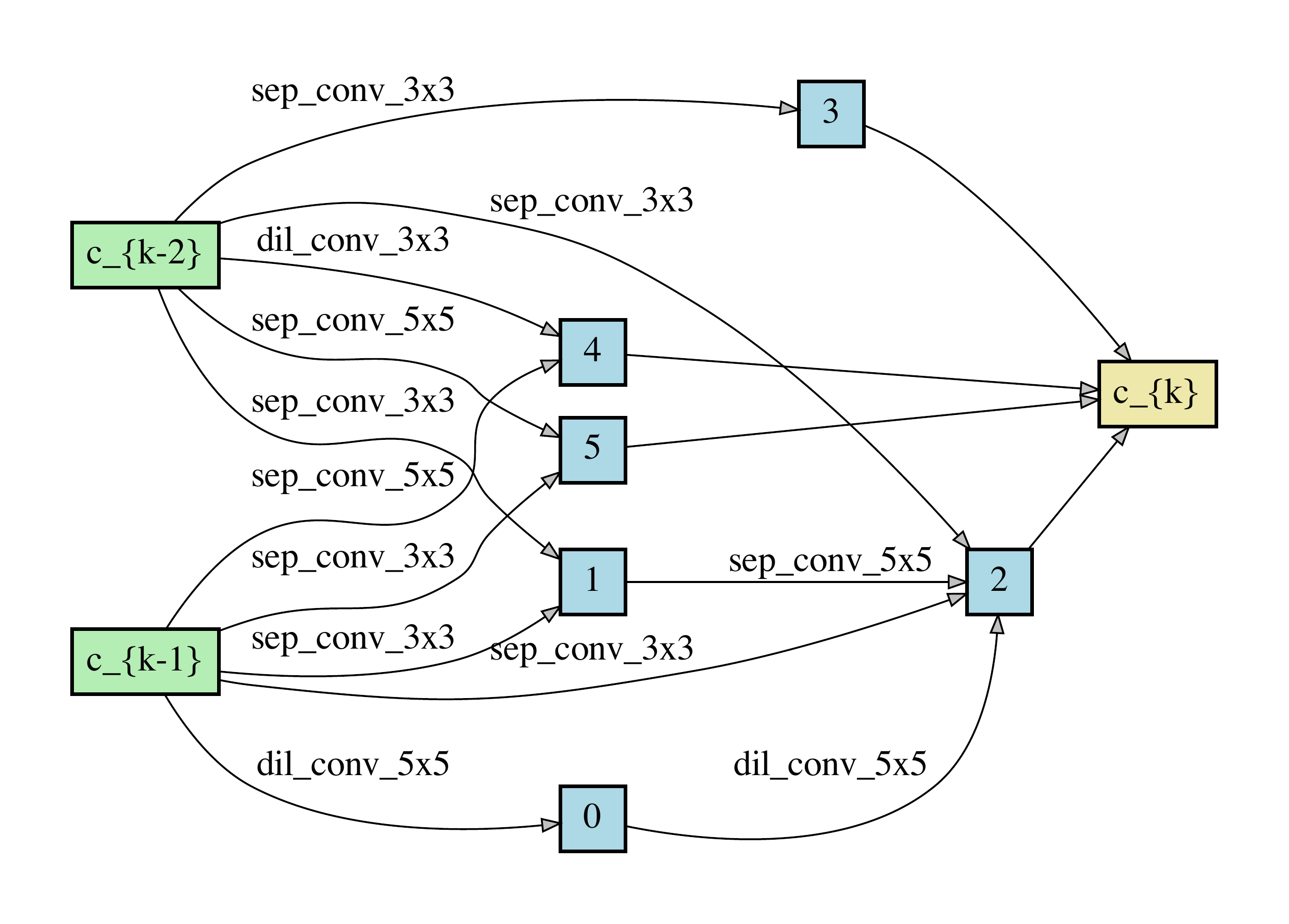}}
	\vspace{-4mm}
	\caption{Visualization of cells obtained by our method.}
	\label{fig:vis}
	\vspace{-6mm}
\end{figure}

We also visualize our obtained normal cells \textit{MiLeNAS-SC-L8} and \textit{MiLeNAS-SC-L16} in Figure~\ref{fig:vis}. We can see that the cell searched with $8$ layers is deeper than the one with $16$ layers, demonstrating that our stretchable cells can learn a suitable depth according to the number of layers. Meanwhile, \textit{MiLeNAS-SC-L8} prefers more dilated convolutions in order to increase the reception field.
More searched cells are visualized in supplementary materials.

\subsection{Results on ImageNet dataset}
We transfer our best cells \textit{MiLeNAS-SC-L12} to ImageNet with the same stacked-layer number $14$ in the evaluation network as DARTS. We also conduct experiments to directly search the architectures on ImageNet using the same optimization settings as \textit{MiLeNAS-SC-L8}. For search cost consideration, the number of nodes is set to $7$, which is the same as DARTS. We sample 3\% data from ImageNet training data and split them into three sets $\D_{tr}$, $\D_{val_\alpha}$ and $\D_{val_\beta}$ with equal size. As illustrated in Table \ref{tab:imagenet}, the obtained architecture, marked as \textit{MiLeNAS-SC-L8 (ImageNet)}, has higher accuracy compared to the architecture transferred from CIFAR-10 and significantly outperforms our baseline methods. Note that we just simply replace CIFAR-10 dataset with ImageNet, by integrating our stretchable cells into other variants such as PC-DARTS \cite{pcdarts} and using more training data, the performance can be further improved.

\begin{table*}
	\renewcommand\arraystretch{1.17}
	\setlength\tabcolsep{1.2mm}
	\centering
	\caption{Evaluation results on CIFAR-10 with consistent numbers of layers.} \vspace{-2mm}
	\label{tab:consistent_layers}
	\footnotesize
	\begin{tabular}{l|c|cc|cc|cc|cc}
		\hline
		\multirow{4}*{Methods} & \multirow{4}*{Number of Nodes} & \multicolumn{8}{c}{Number of Layers} \\
		\cline{3-10}
		~&~& \multicolumn{2}{c}{8}  & \multicolumn{2}{|c}{12} & \multicolumn{2}{|c}{16} & \multicolumn{2}{|c}{20} \\
		\cline{3-10}
		~&~  & Test Err.  & Params & Test Err.  & Params & Test Err.  & Params & Test Err.  & Params \\
		~&~ & (\%) & (M) & (\%) & (M) & (\%) & (M) & (\%) & (M) \\
		\hline
		DARTS (2nd order) & 7 & 3.06 & 3.3 & 2.86 & 3.4 & 2.97 & 3.3 & 2.79 & 3.5 \\ 
		DARTS (2nd order) & 9 & 3.12 & 3.5 & 3.03 & 3.4 & 2.85 & 3.5 & 2.96 & 3.4 \\ 
		MiLeNAS-SC & 9 & 2.58 & 3.5 & 2.36 & 3.5 & 2.51 & 3.6 & 2.54 & 3.5 \\ 
		\hline
	\end{tabular}
\end{table*}
\begin{table*}[htbp]
	\renewcommand\arraystretch{1.17}
	\setlength\tabcolsep{2.1mm}
	\centering
	\footnotesize
	\caption{Evaluation results on CIFAR-10 at different epochs in search.}
	\label{tab:stability}
	\vspace{-2mm}
	\begin{tabular}{l|cc|cc|cc|cc}
		\hline
		\multirow{4}*{Methods} & \multicolumn{8}{c}{Epochs} \\
		\cline{2-9}
		~ &\multicolumn{2}{c|}{20}  & \multicolumn{2}{c}{30} & \multicolumn{2}{|c}{40} & \multicolumn{2}{|c}{50} \\
		\cline{2-9}
		~  & Test Err.  & Params & Test Err.  & Params & Test Err.  & Params & Test Err.  & Params \\
		~ & (\%) & (M) & (\%) & (M) & (\%) & (M) & (\%) & (M) \\
		\hline
		DARTS (2nd order)& 2.93 & 3.6  & \textbf{2.88} & 3.2  & 3.02 & 2.5 & 2.90 & 2.3 \\ 
		DARTS (2nd order)& 3.05 & 3.1 & 3.16 & 2.3 & \textbf{3.00} & 2.3 & 3.41 & 2.1 \\ 
		DARTS (2nd order)& 2.83 & 4.1  & \textbf{2.82} & 3.0  & 2.87 & 2.6 & 2.87 & 2.6 \\ 
		DARTS w/o SC& 2.89 & 2.0  & 2.94 & 1.9  & 2.81 & 1.8 & \textbf{2.71} & 1.8\\
		DARTS w/o SC& 2.91 & 2.0  & 2.89 & 1.8  & 2.83 & 1.8 & \textbf{2.69} & 1.9\\
		DARTS w/o SC& 2.85 & 2.1  & 2.81 & 1.9  & \textbf{2.73} & 1.8 & 2.75 & 1.9\\
		\hline
	\end{tabular}
	\vspace{-4mm}
\end{table*}

\subsection{Ablation Studies}
\noindent\textbf{Using consistent layer number in DARTS\ } 
Our method uses the same layer numbers in search and evaluation. In contrast, DARTS searches on a supernet with $8$ layers and evaluates with $20$ layers. To compare DARTS with our arbitrary-edge search, we conduct experiments to search on multiple layer numbers and then use the same layer numbers for evaluation. The number of nodes in DARTS is set to $7$ or $9$. The results summarized in Table~\ref{tab:consistent_layers} show that DARTS performs poorly in small numbers of layers (i.e., $8$ and $12$), since the searched cells are usually shallow and wide. In contrast, our stretchable cells adapts each layer number well, by using the consistent number of layers, it can obtain promising performance. Note that we keep similar parameters of networks in evaluation for a fair comparison by controlling the initial channel number $C$.

\bigbreak
\noindent\textbf{Effect of numbers of layers\ } 
To verify that our searched cells are best suitable for the corresponding layer numbers, we evaluate them with multiple numbers of layers, and the results are summarized in Figure~\ref{fig:layers}. We can see that our obtained cells achieve the highest accuracies with their corresponding layer numbers, and the best layer number for DARTS cells is $20$ (DARTS searches with $8$ layers), indicating that our searched cells tie closely with the number of layers in search and learn a suitable depth for each number. 

\begin{figure}[h]
	\centering
	\includegraphics[width=0.75\linewidth]{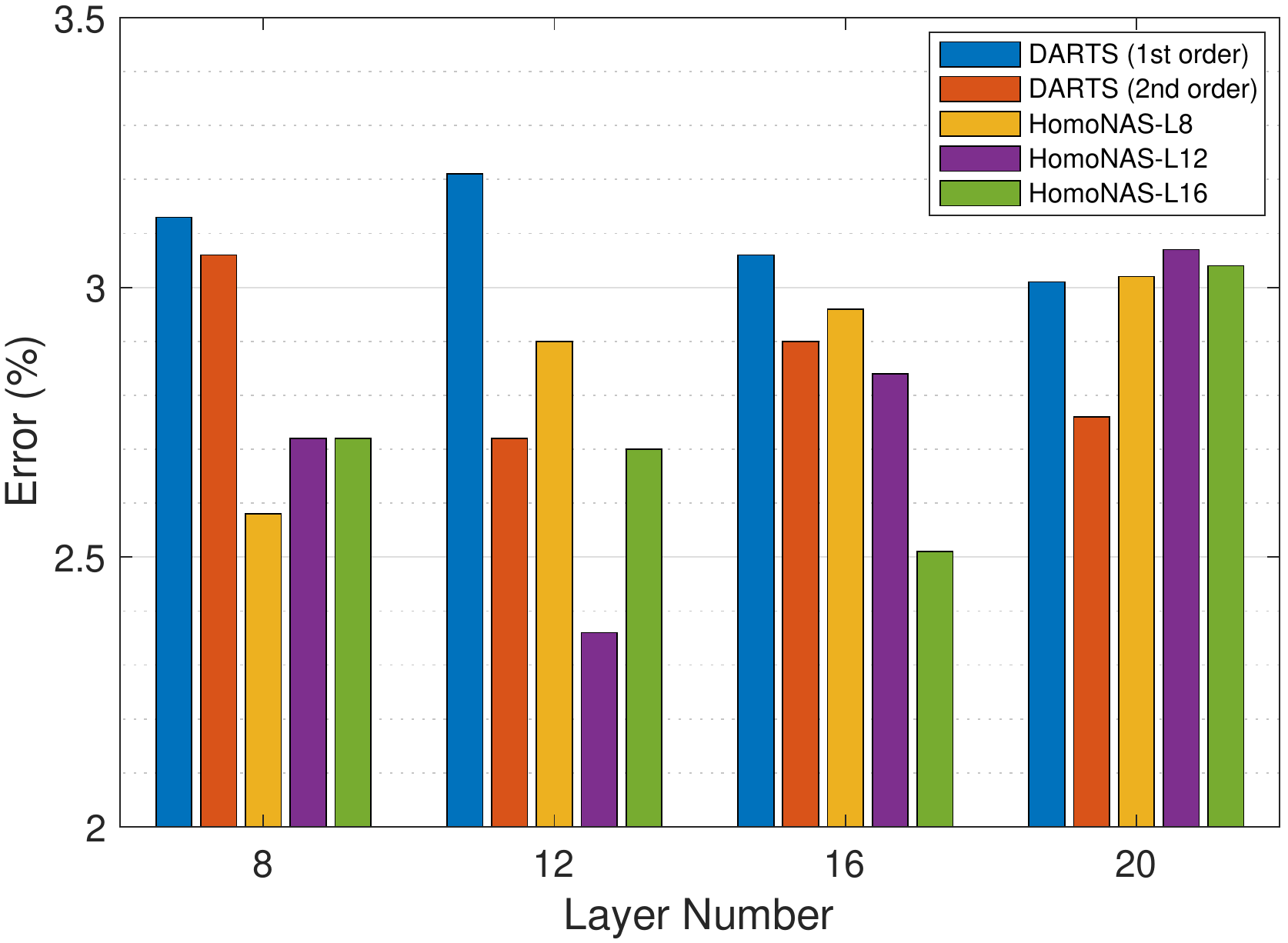}
	\caption{Results with different numbers of layers.}
	\label{fig:layers}
	\vspace{-4mm}
\end{figure}

\bigbreak
\noindent\textbf{Effect of different perspectives of edge selection\ }
As previously discussed, we empirically find that the topology modeling of selecting input edges performs poorly and propose to switch topology space with output edges. In this section, we implement experiments on CIFAR-10 dataset and compare the two manners. As shown in Table~\ref{tb:switch_topo}, modeling topological probabilities with output nodes performs better than using input nodes, and the input manner even performs worse than the original DARTS, which indicates that a lousy topology does harm to the performance of networks.
On the other hand, we switches the input edge selection to output edge selection, making the search space slightly different from DARTS. 
In Table~\ref{tb:switch_topo}, we also investigate the influence of switching input edges to output edges in DARTS. From the results, we can infer that DARTS obtains similar performances on these two perspectives since they have the same edge number (\ie, operation number). Besides, with the same search space of output perspective, our method can still outperform DARTS, which indicates the effectiveness of our explicit topology learning.

\begin{table}[h]
	\centering
	\small
	\caption{Test errors with different perspectives of edge selection.}  
	\label{tb:switch_topo}
	\begin{tabular}{l|c|c}
		\hline
		\multirow{2}*{Methods} & Input Edges & Output Edges \\	
      ~ & (\%) & (\%) \\
		\hline	
		DARTS (2nd order) & 2.76 & 2.82\\
		DARTS w/o SC & 3.08 & 2.72 \\
		\hline
	\end{tabular}
\end{table}

\begin{figure}[h]
	\centering
	\includegraphics[width=0.65\linewidth,height=0.2\textwidth]{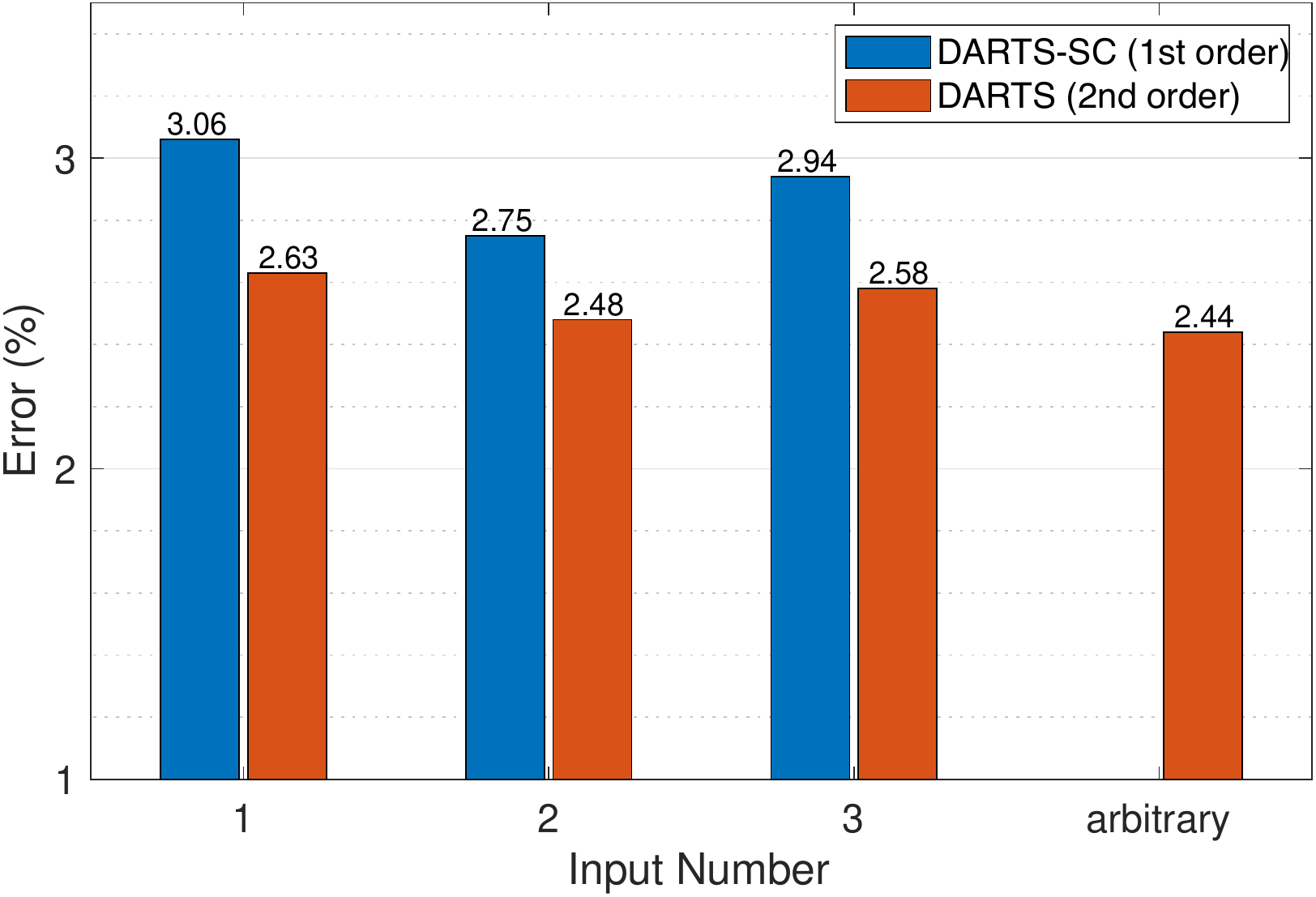}
	\vspace{-2mm}
	\caption{Results with different input numbers.}
	\label{fig:fixed_input_number}
\end{figure}

\noindent\textbf{Comparison between different input numbers of intermediate nodes\ }
DARTS manually chooses two input edges for each intermediate node, while our method can learn arbitrary connections of each node. To investigate the importance of arbitrary connection, we further implement experiments on different but fixed input numbers of each node. Both DARTS and our method use the same number of nodes $N=7$; besides, we use the first-order approximation in our method and second-order approximation in DARTS, marked as \textit{DARTS-SC (1st order)} and \textit{DARTS (2nd order)}, respectively.
From the results shown in Figure~\ref{fig:fixed_input_number}, we can infer that, at the same level of parameters, the increase of input node number does not always ensure the performance improvement. Although more nodes will increase network diversity, since we control a similar amount of parameters, the parameters for each operation will reduce accordingly, limiting their modeling capacity. We can see that input number 2 achieves higher performances since it reaches a better trade-off between cell diversity and modeling capacity per operation. Meanwhile, the arbitrary edge selection performs better than all the fixed edge selections because the network can learn the topology more adaptively instead of being specified by a manual rule. 

\noindent\textbf{Stability during search\ }
DARTS has a severe collapse problem that the \textit{skip-connection} operation will be dominant at the end of search. As previously discussed, we think one reason is that DARTS couples operation and topology together and introduces a \textit{zero} operation for cutting edges, the large weights of \textit{zero} operation would distract the operation selection. In contrast, our stretchable cells decouples the operation and topology then removes the \textit{zero} operation; thus, it should be more stable in search. To this point, we conduct experiments to investigate the stability by evaluating the cells at different epochs in search, and the results are summarized in Table~\ref{tab:stability}. Comparing the performance at different epochs for each run, the accuracies and parameters for DARTS change rapidly during search, and the best architectures often appear before the 50th epoch. However, our method acts more steadily. Note that the best performance in different runs for DARTS also changes more dramatically than our method. In this way, the performance in Table \ref{tab:stability} indicates that our method can obtain a more stable performance during search. We directly use the latest optimized supernet at the 50th epoch for architecture derivation.

\section{Conclusion}
In addition to operations, topology also matters for neural architectures since it controls the interactions between features of operations. In this paper, for a homogenous search of architectures, we highlight the topology learning in differentiable NAS and propose an topology modeling method, which learns stretchable cells w.r.t the network depth. Concretely, we introduce a combinatorial probabilistic distribution to model the target topology explicitly, and leverage a passive-aggressive regularization to suppress invalid topology during search. We apply our stretchable cells to two typical algorithms DARTS and MiLeNAS. Experimental results show that our method can significantly boost the performance on various layer numbers, with better classification accuracy but much fewer parameters. 

\bibliography{main}

\begin{thebibliography}{26}
\providecommand{\natexlab}[1]{#1}

\bibitem[{Anandalingam and Friesz(1992)}]{anandalingam1992hierarchical}
Anandalingam, G.; and Friesz, T.~L. 1992.
\newblock Hierarchical optimization: An introduction.
\newblock \emph{Annals of Operations Research}, 34(1): 1--11.

\bibitem[{Baker et~al.(2017)Baker, Gupta, Naik, and Raskar}]{BakerGNR17}
Baker, B.; Gupta, O.; Naik, N.; and Raskar, R. 2017.
\newblock Designing Neural Network Architectures using Reinforcement Learning.
\newblock In \emph{5th International Conference on Learning Representations,
  {ICLR} 2017, Toulon, France, April 24-26, 2017, Conference Track
  Proceedings}.

\bibitem[{Bender et~al.(2018)Bender, Kindermans, Zoph, Vasudevan, and
  Le}]{bender2018understanding}
Bender, G.; Kindermans, P.-J.; Zoph, B.; Vasudevan, V.; and Le, Q. 2018.
\newblock Understanding and Simplifying One-Shot Architecture Search.
\newblock In \emph{International Conference on Machine Learning}, 550--559.

\bibitem[{Bi et~al.(2019)Bi, Hu, Xie, Chen, Wei, and Tian}]{bi2019stabilizing}
Bi, K.; Hu, C.; Xie, L.; Chen, X.; Wei, L.; and Tian, Q. 2019.
\newblock Stabilizing DARTS with Amended Gradient Estimation on Architectural
  Parameters.
\newblock arXiv:1910.11831.

\bibitem[{Bi et~al.(2020)Bi, Xie, Chen, Wei, and Tian}]{bi2020gold}
Bi, K.; Xie, L.; Chen, X.; Wei, L.; and Tian, Q. 2020.
\newblock Gold-nas: Gradual, one-level, differentiable.
\newblock \emph{arXiv preprint arXiv:2007.03331}.

\bibitem[{Cai, Zhu, and Han(2018)}]{proxylessnas}
Cai, H.; Zhu, L.; and Han, S. 2018.
\newblock ProxylessNAS: Direct Neural Architecture Search on Target Task and
  Hardware.
\newblock In \emph{International Conference on Learning Representations}.

\bibitem[{Chen et~al.(2019)Chen, Xie, Wu, and Tian}]{chen2019progressive}
Chen, X.; Xie, L.; Wu, J.; and Tian, Q. 2019.
\newblock Progressive differentiable architecture search: Bridging the depth
  gap between search and evaluation.
\newblock In \emph{Proceedings of the IEEE/CVF International Conference on
  Computer Vision}, 1294--1303.

\bibitem[{Colson, Marcotte, and Savard(2007)}]{colson2007overview}
Colson, B.; Marcotte, P.; and Savard, G. 2007.
\newblock An overview of bilevel optimization.
\newblock \emph{Annals of operations research}, 153(1): 235--256.

\bibitem[{Deng et~al.(2009)Deng, Dong, Socher, Li, Li, and Fei-Fei}]{Imagenet}
Deng, J.; Dong, W.; Socher, R.; Li, L.-J.; Li, K.; and Fei-Fei, L. 2009.
\newblock Imagenet: A large-scale hierarchical image database.
\newblock In \emph{2009 IEEE conference on computer vision and pattern
  recognition}, 248--255. Ieee.

\bibitem[{He et~al.(2020)He, Ye, Shen, and Zhang}]{milenas}
He, C.; Ye, H.; Shen, L.; and Zhang, T. 2020.
\newblock Milenas: Efficient neural architecture search via mixed-level
  reformulation.
\newblock In \emph{The IEEE Conference on Computer Vision and Pattern
  Recognition (CVPR)}.

\bibitem[{Howard et~al.(2017)Howard, Zhu, Chen, Kalenichenko, Wang, Weyand,
  Andreetto, and Adam}]{howard2017mobilenets}
Howard, A.~G.; Zhu, M.; Chen, B.; Kalenichenko, D.; Wang, W.; Weyand, T.;
  Andreetto, M.; and Adam, H. 2017.
\newblock Mobilenets: Efficient convolutional neural networks for mobile vision
  applications.
\newblock \emph{arXiv preprint arXiv:1704.04861}.

\bibitem[{Huang et~al.(2017)Huang, Liu, Van Der~Maaten, and
  Weinberger}]{huang2017densely}
Huang, G.; Liu, Z.; Van Der~Maaten, L.; and Weinberger, K.~Q. 2017.
\newblock Densely connected convolutional networks.
\newblock In \emph{Proceedings of the IEEE conference on computer vision and
  pattern recognition}, 4700--4708.

\bibitem[{Krizhevsky, Nair, and Hinton(2014)}]{cifar10}
Krizhevsky, A.; Nair, V.; and Hinton, G. 2014.
\newblock The cifar-10 dataset.
\newblock \emph{online: http://www. cs. toronto. edu/kriz/cifar. html}, 55.

\bibitem[{Liang et~al.(2019)Liang, Zhang, Sun, He, Huang, Zhuang, and
  Li}]{liang2019darts}
Liang, H.; Zhang, S.; Sun, J.; He, X.; Huang, W.; Zhuang, K.; and Li, Z. 2019.
\newblock DARTS+: Improved Differentiable Architecture Search with Early
  Stopping.
\newblock arXiv:1909.06035.

\bibitem[{Liu, Simonyan, and Yang(2018)}]{darts}
Liu, H.; Simonyan, K.; and Yang, Y. 2018.
\newblock DARTS: Differentiable Architecture Search.
\newblock In \emph{International Conference on Learning Representations}.

\bibitem[{Ma et~al.(2018)Ma, Zhang, Zheng, and Sun}]{ma2018shufflenet}
Ma, N.; Zhang, X.; Zheng, H.-T.; and Sun, J. 2018.
\newblock Shufflenet v2: Practical guidelines for efficient cnn architecture
  design.
\newblock In \emph{Proceedings of the European Conference on Computer Vision
  (ECCV)}, 116--131.

\bibitem[{Pham et~al.(2018)Pham, Guan, Zoph, Le, and Dean}]{pham2018efficient}
Pham, H.; Guan, M.; Zoph, B.; Le, Q.; and Dean, J. 2018.
\newblock Efficient Neural Architecture Search via Parameters Sharing.
\newblock In \emph{International Conference on Machine Learning}, 4095--4104.

\bibitem[{Real et~al.(2019)Real, Aggarwal, Huang, and Le}]{real2019regularized}
Real, E.; Aggarwal, A.; Huang, Y.; and Le, Q.~V. 2019.
\newblock Regularized evolution for image classifier architecture search.
\newblock In \emph{Proceedings of the aaai conference on artificial
  intelligence}, volume~33, 4780--4789.

\bibitem[{Shu, Wang, and Cai(2019)}]{shu2019understanding}
Shu, Y.; Wang, W.; and Cai, S. 2019.
\newblock Understanding architectures learnt by cell-based neural architecture
  search.
\newblock \emph{arXiv preprint arXiv:1909.09569}.

\bibitem[{Tan et~al.(2019)Tan, Chen, Pang, Vasudevan, Sandler, Howard, and
  Le}]{tan2019mnasnet}
Tan, M.; Chen, B.; Pang, R.; Vasudevan, V.; Sandler, M.; Howard, A.; and Le,
  Q.~V. 2019.
\newblock Mnasnet: Platform-aware neural architecture search for mobile.
\newblock In \emph{Proceedings of the IEEE Conference on Computer Vision and
  Pattern Recognition}, 2820--2828.

\bibitem[{Wang et~al.(2020)Wang, Dai, Li, Zou, and Xiong}]{wang2020si}
Wang, Y.; Dai, W.; Li, C.; Zou, J.; and Xiong, H. 2020.
\newblock Si-vdnas: Semiimplicit variational dropout for hierarchical one-shot
  neural architecture search.
\newblock In \emph{International Joint Conference on Artificial Intelligence}.

\bibitem[{Xie et~al.(2019)Xie, Zheng, Liu, and Lin}]{snas}
Xie, S.; Zheng, H.; Liu, C.; and Lin, L. 2019.
\newblock {SNAS:} stochastic neural architecture search.
\newblock In \emph{7th International Conference on Learning Representations,
  {ICLR} 2019, New Orleans, LA, USA, May 6-9, 2019}.

\bibitem[{Xu et~al.(2019)Xu, Xie, Zhang, Chen, Qi, Tian, and Xiong}]{pcdarts}
Xu, Y.; Xie, L.; Zhang, X.; Chen, X.; Qi, G.-J.; Tian, Q.; and Xiong, H. 2019.
\newblock PC-DARTS: Partial Channel Connections for Memory-Efficient
  Architecture Search.
\newblock In \emph{International Conference on Learning Representations}.

\bibitem[{Yan et~al.(2019)Yan, Fang, Zhang, Zheng, Zeng, Zhang, and
  Xu}]{yan2019hm}
Yan, S.; Fang, B.; Zhang, F.; Zheng, Y.; Zeng, X.; Zhang, M.; and Xu, H. 2019.
\newblock Hm-nas: Efficient neural architecture search via hierarchical
  masking.
\newblock In \emph{Proceedings of the IEEE/CVF International Conference on
  Computer Vision Workshops}, 0--0.

\bibitem[{Zoph and Le(2017)}]{ZophL17}
Zoph, B.; and Le, Q.~V. 2017.
\newblock Neural Architecture Search with Reinforcement Learning.
\newblock In \emph{5th International Conference on Learning Representations,
  {ICLR} 2017, Toulon, France, April 24-26, 2017, Conference Track
  Proceedings}.

\bibitem[{Zoph et~al.(2018)Zoph, Vasudevan, Shlens, and Le}]{nasnet}
Zoph, B.; Vasudevan, V.; Shlens, J.; and Le, Q.~V. 2018.
\newblock Learning transferable architectures for scalable image recognition.
\newblock In \emph{Proceedings of the IEEE conference on computer vision and
  pattern recognition}, 8697--8710.

\end{thebibliography}

\end{document}